\documentclass[letterpaper, 10 pt, conference]{ieeeconf}  
\usepackage[utf8]{inputenc}
\IEEEoverridecommandlockouts                              
\usepackage[nolist,nohyperlinks]{acronym}
\usepackage{glossaries}
\usepackage{cite}
\usepackage{orcidlink}
\usepackage{tabularx}
\usepackage{pifont}
\usepackage{graphicx}
\usepackage{pstricks}
\usepackage{verbatim}
\usepackage{psfrag}
\usepackage[english]{babel}
\usepackage{amsmath}
\usepackage{amssymb}
\usepackage[font=small]{caption}
\usepackage{fancybox}
\usepackage{steinmetz}
\usepackage{subcaption}
\usepackage{bodegraph}
\usepackage{gensymb}
\usepackage{algorithm,algorithmic}
\usepackage{amsfonts}
\usepackage{dsfont}
\usepackage{siunitx}
\usepackage{lscape}
\usepackage{enumerate}
\usepackage{lmodern}
\usepackage[T1]{fontenc}
\usepackage[framed]{matlab-prettifier}
\usepackage{booktabs}
\usepackage{mathtools}
\usepackage{color} 
\usepackage{xcolor,colortbl}
\usepackage{colortbl}
\usepackage{multirow, hhline}
\usepackage{url}
\usepackage{bm}
\usepackage{multicol}
\usepackage{supertabular}
\usepackage{graphicx}
\usepackage{hyperref}
\usepackage{xcolor}
\usepackage{eso-pic}

\definecolor{verdemio}{rgb}{0.01, 0.75, 0.24}

\makeatletter
\newcommand\notsotiny{\@setfontsize\notsotiny\@vipt\@viipt}
\makeatother
\graphicspath{{Images/}}
\newacronym{fsm}{FSM}{Finite State Machine}
\newacronym{slerp}{SLERP}{Spherical Linear Interpolation}
\newacronym{vlms}{VLMs}{Vision-Language Models}
\newacronym{llms}{LLMs}{Large Language Models}
\newacronym{sus}{SUS}{System Usability Scale}
\newacronym{pHRI}{pHRI}{physical Human-Robot Interaction}
\definecolor{Gray}{gray}{0.9}

\definecolor{verdedd}{rgb}{0.0, 0.5, 0.0}

\definecolor{myorangenew}{rgb}{1 0.49 0}
\definecolor{mygreennew}{rgb}{0.31 0.78 0.47}
\definecolor{mygreynew}{rgb}{0.6 0.6 0.6}
\definecolor{bluD}{rgb}{0.09, 0.45, 0.81}
\definecolor{bluD_l}{rgb}{0.29, 0.65, 1}
\definecolor{nero}{rgb}{0, 0, 0}
\definecolor{bianco}{rgb}{1, 1, 1}

\definecolor{bluG}{cmyk}{100 85 0 0}
\definecolor{verdeG}{cmyk}{75 0 100 0}

   \def\p{{\boldsymbol p}}

 \def\x{{\boldsymbol x}}

 \def\0{{\boldsymbol 0}}

\DeclareMathSymbol{\boldmu}{\mathord}{letters}{15}
\newcommand{\T}[2]{{}^{\scriptstyle #1}\mathbf{T}_{\scriptstyle #2}}
\newcommand{\R}[2]{{}^{\scriptstyle #1}\mathbf{R}_{\scriptstyle #2}}
\newcommand{\pp}[2]{{}^{\scriptstyle #1}\boldsymbol{p}_{\scriptstyle #2}}

\begin{document}

\AddToShipoutPictureBG*{%
  \AtPageUpperLeft{%
    \put(0,-45){%
      \makebox[\paperwidth][c]{
        \parbox{\textwidth}{
          \centering
          \scriptsize
          \textcopyright~\textit{2026 IEEE. Personal use of this material is permitted. Permission from IEEE must be obtained for all other uses, in any current or future media, including reprinting/republishing this material for advertising or promotional purposes, creating new collective works, for resale or redistribution to servers or lists, or reuse of any copyrighted component of this work in other works.\\[1ex]}
          \textbf{Preprint version.} This work has been accepted for publication at the 35th IEEE International Conference on Robot and Human Interactive Communication (RO-MAN 2026), and should comply with applicable copyright and open-access policies.
        }%
      }%
    }%
  }%
}
\title{\textbf{Adaptive vs. Static Robot-to-Human Handover:}\\\textbf{ A Study on Orientation and Approach Direction}}
\author{Federico Biagi$^{1}$, Dario Onfiani$^{1}$, Simone Silenzi$^{1}$, Cristina Iani$^{2}$, Luigi Biagiotti$^{1}$
\thanks{This work was partially supported by the Italian National Recovery and Resilience Plan (PNRR), Mission 4 “Education and Research”, Component C2, Investment 1.1 “PRIN – Projects of Relevant National Interest”, Project I-SHARM: Intelligent SHared Autonomy for Robotic Manipulation Systems, Project ID 2022NTZRFM, CUP E53C24002600006 and by the University of Modena and Reggio Emilia under the FAR (Fondo di Ateneo per la Ricerca – Linea Fomo) project titled \textit{ROBIN3: a ROBotic INTelligent, INTuitive, and INTeractive platform for NAO-Mediated Autistic Healthcare}.
Furthermore, the authors acknowledge the use of artificial intelligence tools to assist with English language editing and to aid in the generation of illustrative elements within the figures.}
\thanks{$^{1}$ Department of Engineering "Enzo Ferrari", University of Modena and Reggio Emilia, Italy. Emails: \{federico.biagi, dario.onfiani, simone.silenzi, luigi.biagiotti\}@unimore.it}
\thanks{$^{2}$ Department of Surgery, Medicine, Dentistry and Morphological Sciences, University of Modena and Reggio Emilia, Italy. Email: cristina.iani@unimore.it}}

\maketitle
\begin{abstract}
Robot-to-human handovers often rely on static, open-loop strategies (or, at best, approaches that adapt only the position), which generally do not consider how the object will be grasped by the human, thus requiring the user to adapt. This work presents a novel adaptive framework that dynamically adjusts the object's delivery pose in real time based on the user's hand pose and the intended downstream task. By integrating AI-based hand pose estimation with smooth, kinematically constrained trajectories, the system ensures a safe approach and an optimal handover orientation.
A comprehensive user study compares the proposed adaptive approach against a static baseline across multiple tasks, evaluating both subjective metrics (NASA-TLX, Human-Robot Trust Scale) and objective physiological data (blink rate measured via wearable eye-trackers). The results demonstrate that dynamic alignment significantly reduces users' cognitive workload and physiological stress, while improving their confidence in the robot's reliability. These findings highlight the potential of task- and pose-aware systems for enabling fluid and ergonomic human-robot collaboration.
\end{abstract}
\section{Introduction}
\label{sec:Introduction}
Human-robot interaction encompasses a wide range of collaborative tasks that are increasingly integrating service and industrial collaborative robots (cobots) into shared human environments \cite{ajoudani2018progress, elzaatari2019cobot}. Within these collaborative scenarios, \acrfull{pHRI} plays a pivotal role, and the robot-to-human object handover emerges as one of the most fundamental and ubiquitous joint actions \cite{ortenzi2021object}. Unlike conventional pick-and-place operations, a successful handover is a highly coordinated joint action requiring precise spatio-temporal alignment, intention understanding, and minimized receiver effort\cite{strabala2013towards, brand2022predictability}. Recent large-scale studies on human-human handovers highlight that even among humans, misalignments between the giver's presentation and the receiver's grasp intentions frequently occur, leading to a measurable decrease in interaction comfort and fluency \cite{wiederhold2023studying, wiederhold2025studying}.

In seamless human-human interactions, the giver intuitively adapts the object's position and orientation to match the receiver's posture and intended subsequent use \cite{parastegari2017modeling}. This intrinsic understanding ensures that the delivery configuration avoids interference with the human's preferred grasp region, allowing them to transition smoothly into the downstream task without cumbersome regraspings \cite{brand2022predictability}. In fact, humans naturally adopt \textit{receiver-centered} strategies, actively aligning the object's affordance axes (e.g., presenting the handle of a tool) to accommodate the partner's needs \cite{chan2015characterization}. Conversely, standard robot-to-human handovers often rely on static, open-loop strategies \cite{chan2013human}. Typically, the robot transports the object to a predefined spatial coordinate and rigidly waits for the human to retrieve it. This object-centric approach forces the human to adapt to the robot's configuration, which, as demonstrated by the aforementioned intention misalignment studies, often results in awkward grasping postures, increased physical strain, and a suboptimal user experience \cite{cakmak2011human}.

\begin{figure}[t]
    \vspace{+2.5mm}
    \centering
    \begin{subfigure}[b]{0.48\columnwidth}
        \centering
        \includegraphics[width=\textwidth, height=4.5cm]{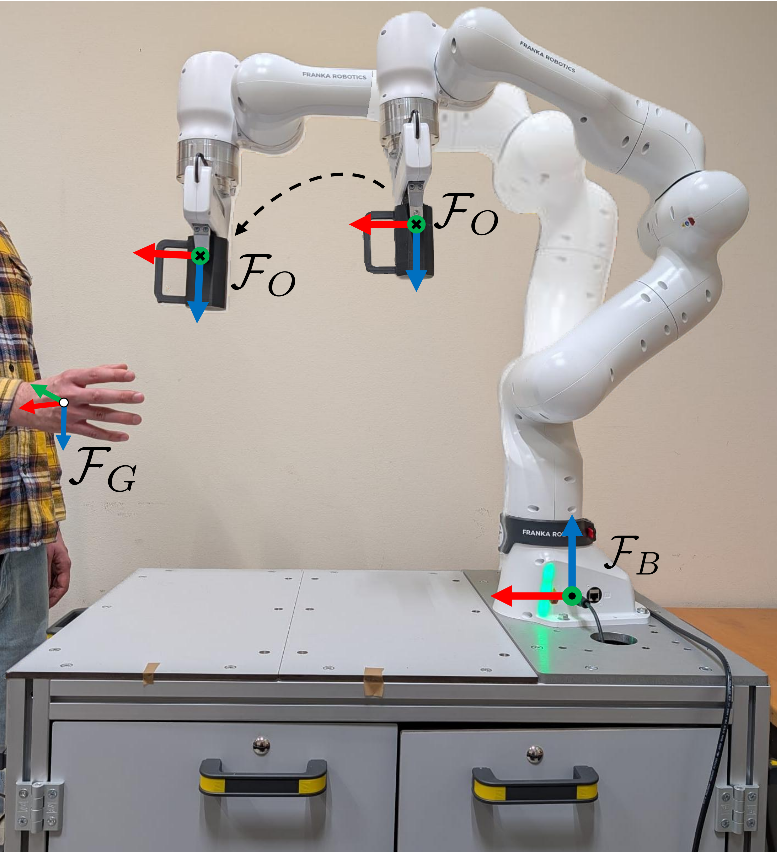}
        \caption{Static Handover}
        \label{fig:image1}
    \end{subfigure}
    \hfill
    \begin{subfigure}[b]{0.48\columnwidth}
        \centering
        \includegraphics[width=\textwidth, height=4.5cm]{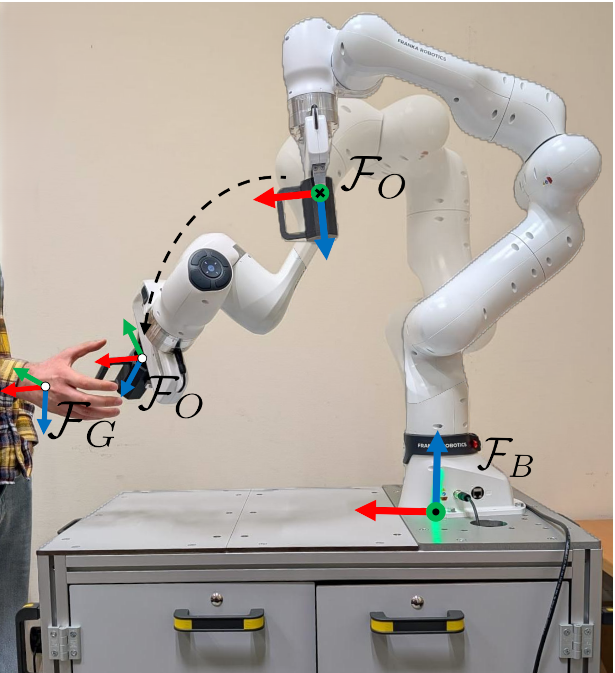}
        \caption{Adaptive Handover}
        \label{fig:image2}
    \end{subfigure}
    \caption{Conceptual comparison between traditional static robot-to-human handovers (a) and our proposed adaptive, task-oriented approach (b).}
    \label{fig:intro_concept}
    \vspace{-5mm}
    \label{fig:intro}
\end{figure}

%
To achieve truly fluid collaboration, robotic systems must transition from rigid executors to adaptive partners \cite{chan2013human}. This entails dynamically adjusting the handover location based on the human's hand pose and aligning the object's orientation according to the affordances required by the downstream task (Fig. \ref{fig:intro}).
Applications extend beyond able-bodied users retrieving items, targeting collaborative settings like industrial assistance, rehabilitation, eldercare, or domestic support.
However, designing such reactive systems introduces new challenges: continuous kinematic adaptation can make the robot's motion unpredictable, potentially raising the user's cognitive workload and reducing perceived safety. Therefore, it is crucial not only to develop advanced, human-centric kinematic control strategies that fuse real-time tracking with geometric fluidity, but also to rigorously evaluate their impact on the user's psychological and physiological state during the interaction \cite{dehais2011physiological}.

\section{Related Works}
\label{sec:RelatedWorks}
%
\begin{table}[t]
\centering
\caption{Comparison of robot-to-human handover methods}
\label{tab:sota_comparison}
\footnotesize
\setlength{\tabcolsep}{4pt}
\renewcommand{\arraystretch}{1.1}
\resizebox{\columnwidth}{!}{
\begin{tabular}{@{}l p{1.6cm} p{1.8cm} p{1.6cm} p{1.6cm}@{}}
\toprule
\textbf{Reference} 
& \textbf{Real-Time Tracking} 
& \textbf{Task-Aware Alignment} 
& \textbf{Predictable Motion} 
& \textbf{Objective Metrics} \\ \midrule
\cite{aleotti2012comfortable} & No & Partial & Yes & No \\
\cite{liu2025leveraging}, \cite{tulbure2025llm} & No & Yes & Yes & No \\
\cite{zhang2023flexible} & Yes & No & No & No \\
\cite{kappler2023optimizing}, \cite{sorrentino2024investigating} & Yes & No & No & No \\ \midrule
\textbf{Ours} & \textbf{Yes} & \textbf{Yes} & \textbf{Yes} & \textbf{Yes} \\ \bottomrule
\end{tabular}
}
\end{table}

%
The topic of robot-to-human handover has seen a transition from static transfer strategies to highly adaptive and context-aware behaviors. A common approach to improve ergonomics involves planning the robot's pose so that the graspable parts of an object are directed toward the user \cite{aleotti2012comfortable}. Recent zero-shot approaches employ \acrfull{vlms} to extract semantic and geometric information, computing optimal delivery configurations prior to motion execution \cite{liu2025leveraging}. Furthermore, task-oriented handovers deduce the optimal orientation from human demonstrations \cite{qin2022task}, and very recently, \acrfull{llms} have been successfully exploited to reason about task constraints, allowing the robot to present tools optimally tailored for the subsequent human action \cite{tulbure2025llm}. However, these approaches are predominantly object-centric and static; they compute a spatial target pose a priori and maintain it fixed during execution, failing to adapt the object's orientation in real-time to continuous changes in the human wrist posture during the approach phase.
The necessity to handle continuously moving users has led to the development of robust adaptive trajectories for real-time tracking \cite{zhang2023flexible}. Nevertheless, continuous tracking presents a known critical issue: continuously updating the target position based on the human hand can make the robot's movements unpredictable. A comprehensive study presented in \cite{kappler2023optimizing} demonstrated that adaptive transport methods, which continuously track the hand, can significantly reduce the user's trust and perceived safety compared to predefined static trajectories, due to the higher cognitive effort required by the human receiver to anticipate the robot's intentions and path changes. Similar dynamics of cognitive overload and behavioral habituation in repetitive handover scenarios have also been highlighted by monitoring the psychological engagement of users \cite{sorrentino2024investigating}. \\
The system proposed in our work bridges these critical gaps in the literature. As summarized in Table \ref{tab:sota_comparison}, existing approaches typically exhibit a trade-off among four fundamental features for seamless human-robot interaction: real-time spatial tracking, task-aware object alignment, kinematic predictability, and the use of objective physiological metrics. 
Unlike static methods that can ensure predictability and task-awareness but completely ignore user movement during the approach \cite{aleotti2012comfortable, liu2025leveraging, tulbure2025llm}, or dynamic tracking methods that adapt in real-time but degrade user trust due to erratic, unpredictable motions \cite{zhang2023flexible, kappler2023optimizing}, our framework addresses this conflict by geometrically decoupling the spatial approach, governed by a highly predictable, limited-jerk path, from the continuous task-oriented alignment of the object. Furthermore, while previous comparative studies \cite{kappler2023optimizing, sorrentino2024investigating} primarily rely on subjective feedback to assess human-robot trust and cognitive load, we introduce a multimodal evaluation framework grounded in objective physiological data. Consequently, our proposed framework provides an initial integrated approach that combines these interaction constraints within a single handover framework. The main contributions of our work are twofold:

\begin{itemize}
    \item \textit{Dynamic Kinematic Control:} The system simultaneously considers the optimal position and orientation relative to the user's hand in real-time. By utilizing AI pose estimation models, we extract the kinematics of the human wrist. To ensure a fluid and predictable approach that preserves user trust, we generate Cartesian trajectories on B\'ezier curves geometrically constrained by the optimal approach direction toward the user's palm, guaranteeing a natural and unobstructed presentation of the object. The translational motion is parameterized by a fifth-order polynomial time law designed to limit the maximum jerk. Through SLERP interpolation, the object's attitude is adapted on the fly, dynamically aligning the object based on the specific downstream task the user will perform \footnote{Link to the project website: \url{https://autolabmodena.github.io/adaptive-human-handover/}}.
    \item \textit{Comparative Experimental Validation:} To assess the actual impact of our adaptive approach on cognitive workload and user trust compared to a static baseline, we conducted a comprehensive user study. This validation combines subjective metrics via validated questionnaires, such as the NASA Task Load Index (NASA-TLX) \cite{hart2006} and a modified Human-Robot Trust Scale  \cite{Charalambous_2015} adapted to our case study, with objective physiological data acquired through NEON Pupil Labs glasses wearable eye-trackers. The real-time analysis of visual parameters, specifically the blink rate, allows us to quantify the user's cognitive effort \cite{Ifrah_2025}, offering a holistic and robust evaluation of the handover quality.
\end{itemize}

\section{Study Description}
\label{sec.Study_Description}
\begin{figure*}[t]
    \centering
    \includegraphics[width=\textwidth]{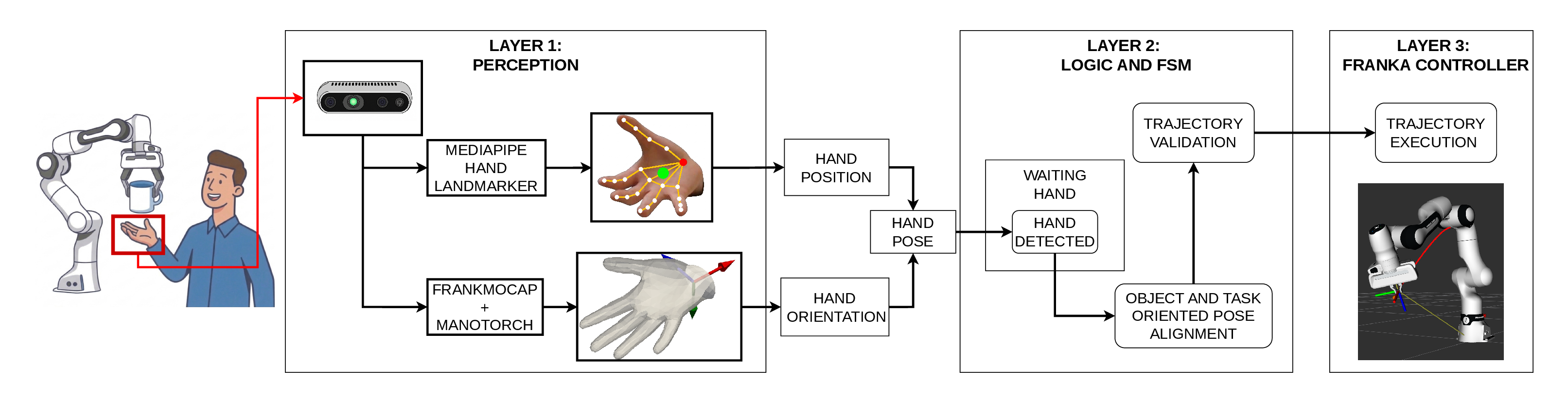}
    \caption{Overview of the proposed Adaptive handover framework architecture. The system is divided into three main layers: Layer 1 extracts 3D hand position and orientation using MediaPipe and FrankMocap + Manotorch; Layer 2 orchestrates the handover Finite State Machine (FSM) and computes the object-task-oriented pose alignment while also validating the trajectory; Layer 3 executes the safe trajectory.}
    \label{fig:architecture_framework}
    \vspace{-3mm}
\end{figure*}
A user study was conducted comparing the \textit{Adaptive Handover} against a \textit{Static Handover} baseline. We hypothesized that by dynamically aligning the object's pose to the user's hand and intended downstream task, the adaptive approach would significantly improve interaction ergonomics while reducing the user's physical effort and cognitive workload compared to the stationary baseline.
\subsection{Participants}
\label{subsec:partecipants}
Fourteen right-handed participants (2 female, 12 male, primarily under 35 years old) volunteered for the study. Half of the sample had prior experience with robotics. The study complied with the Declaration of Helsinki and received institutional approval. All participants provided informed written consent.
\subsection{Apparatus}
\label{subsec.Apparatus}
The experimental architecture integrates perception, orchestration, and low-level control to achieve fluid handovers (Fig.~\ref{fig:architecture_framework}). The setup features a Franka Emika Panda cobot and an Intel RealSense D435i camera. The perception layer fuses 2D keypoints and depth data via MediaPipe Hands \cite{mediapipe_2020} for 3D wrist localization, alongside FrankMocap \cite{rong2021frankmocap} and Manotorch \cite{yang2021cpf} for spatial orientation. These inputs enable a dynamical alignment of the end-effector to the user's hand pose. Finally, the target trajectory is validated through an inverse kinematics solver to prevent singularities and joint limit violations.

\subsection{Kinematics and Hand Target Computation}
The employed architecture fuses distinct data sources. The system defines four primary coordinate frames: the Camera frame ($\mathcal{F}_{C}$), the Grasp frame ($\mathcal{F}_{G}$), the Robot Base frame ($\mathcal{F}_{B}$) and the Hand frame ($\mathcal{F}_{H}$). To represent poses and compute spatial relationships concisely, we adopt homogeneous transformation matrices $\T{}{} \in SE(3)$.
The hand's spatial position $\pp{C}{H} \in \mathbb{R}^{3}$ is obtained by back-projecting the 2D MediaPipe landmarks (specifically, the mean pixel position between the wrist and the middle finger metacarpus) using the camera's aligned depth map. FrankMocap reconstructs the hand shape, outputting the wrist's rotation matrix $\R{C}{H}$ relative to the camera via Manotorch. These components are embedded into a single homogeneous transformation matrix representing the hand pose in the camera frame:
\begin{equation}
    \T{C}{H} = 
    \begin{bmatrix}
        \R{C}{H} & \pp{C}{H} \\
        0_{1 \times 3} & 1
    \end{bmatrix}
\end{equation}
Utilizing a static hand-eye calibration matrix $\T{B}{C}$, which describes the transformation from the camera to the robot base, the hand's spatial pose is transformed into the robot's base frame through a direct matrix multiplication:
\begin{equation}
    \T{B}{H} = \T{B}{C}\T{C}{H}
    \nonumber
\end{equation}
For each object, the grasp reference frame (coinciding with the object frame during handover) is defined with respect to the hand frame to ensure task-dependent ergonomic optimization. This is achieved by combining a displacement along the palm axis with a task-specific orientation.\\
These effects are encoded in a constant homogeneous transformation $\T{H}{G}$ (specific to each scenario), which captures both the ergonomic offset and the desired grasp orientation to ensure a seamless grasp. The parameters of $\T{H}{G}$ were derived from a preliminary user study, in which the participants indicated their preferred receiving orientations for different post-handover tasks. In the present implementation, the downstream task is not inferred autonomously by the robot. Instead, the task label is specified by the experimental protocol for each trial and is used to select the corresponding task-specific transformation $\T{H}{G}$. Therefore, the online adaptation concerns the receiver's measured hand pose.
The optimal grasping pose in the robot base frame, $\T{B}{G}$, is then computed as:
\begin{equation}
    \T{B}{G} = \T{B}{H}\T{H}{G}
    \nonumber
\end{equation}

Finally, the target pose for the object required to execute the handover is obtained as:
\begin{equation}
    \hat{\T{B}{O}} = \T{B}{G}.
    \label{eq:TargetPose}
\end{equation}

The transformation $\T{B}{O}$ includes the forward kinematics of the robot arm and a constant transformation that depends on the relative pose between the gripper and the object.

\subsection{Robot Control and Trajectory Generation}
\label{subsec:control}
\begin{figure}[tb]
    \centering
    \includegraphics[width=0.7\columnwidth]{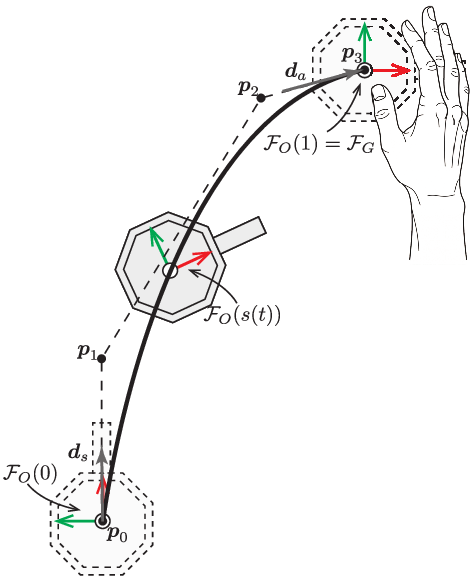}
    \caption{Schematic representation of the cubic Bézier trajectory generated for the handover task of a mug. The geometric path is shaped by four control points ($\p_0$,$\p_1$,$\p_2$,$\p_3$), imposing the start ($\boldsymbol{d}_{s}$) and approach ($\boldsymbol{d}_{a}$) directions to ensure a smooth, natural motion toward the final grasp frame ($\mathcal{F}_{G}$).}
    \label{fig:bezier}
\vspace{-4mm}
\end{figure}

The robot's control architecture is developed in ROS2. The main script evaluates a \acrfull{fsm} at a fixed loop rate to autonomously manage both the handover of objects to the user and the sequential retrieval of new objects for the user study. The system operates in two main modalities:

\begin{enumerate}
    \item \textit{Static Handover:} The robot moves the object to a fixed, predefined pose $\bar{\T{B}{O}}$ in the handover area, completely independent of the user's hand pose or downstream task.
    \item \textit{Adaptive Handover:} The robot computes a dynamic target $\hat{\T{B}{O}}$ from \eqref{eq:TargetPose}, adjusting the delivery pose based on the user hand pose and intended downstream task to ensure optimal affordance.
\end{enumerate} 
The object pose trajectory is defined by decoupling the geometric path (for both position and orientation) from the motion law, i.e.,
\begin{equation}
    \T{B}{O}(s) =
    \begin{bmatrix}
        \R{B}{O}(s) & \pp{B}{O}(s) \\
        \0_{1 \times 3} & 1
    \end{bmatrix}
    \nonumber
\end{equation}
with $s(t) \in [0,1]$.\\
To generate a smooth and ergonomically predictable handover motion, the position $\pp{B}{O}(s)$ is defined as a cubic B\'ezier curve in Cartesian space \cite{biagiotti2008trajectory}:
%
\begin{equation}
    \pp{B}{O}(s) = \sum_{i=0}^{3} \binom{3}{i} (1-s)^{3-i} s^{i} \p_i
    \nonumber
\end{equation}
Here, $\p_0$ and $\p_3$ correspond to the current object position and the final target grasp position, respectively. The intermediate control points shape the approach: $\p_1$ is defined along the initial object $X$-axis ($\x_{O}$) to enforce a smooth tangent departure along the initial direction $\boldsymbol{d}_{s}$, while $\p_2$ is obtained by retracting from $\p_3$ along the desired approach direction $\boldsymbol{d}_{a}$, oriented outward from the user's palm to ensure a natural approach. Formally,
\begin{align}
    \p_1 &= \p_0 + \alpha_{s} \boldsymbol{d}_{s} \nonumber \\
    \p_2 &= \p_3 - \alpha_{a} \boldsymbol{d}_{a} \nonumber
\end{align}
where $\alpha_{s}$ and $\alpha_{a}$ are tunable scaling factors. The resulting position path is illustrated in Fig.~\ref{fig:bezier}.
The object orientation $\R{B}{O}(s)$ is interpolated via \acrfull{slerp} between the initial and target quaternions. To ensure that the orientation stabilizes before reaching the final position, a tunable orientation-lock factor compresses the rotational interpolation within a fraction of the total trajectory duration. This guarantees that the object reaches the desired task-oriented attitude as it approaches the user's hand.
Finally, the motion law along the geometric path is defined by a fifth-order polynomial $s(t)$, designed to limit the maximum jerk and ensure a smooth and comfortable motion for the human receiver.\\
Prior to execution, a Pinocchio-based inverse kinematics solver \cite{carpentier2019pinocchio} validates the planned trajectory against manipulability thresholds, singular values, and joint-limit constraints. If any safety criterion is violated, the system triggers an adaptive scaling, iteratively reducing the factor $\alpha$ to shrink the intermediate control points $\p_1$ and $\p_2$. This validation loop guarantees a safe, feasible path without interrupting the handover flow.
\subsection{Experimental Setup and Task Design}
\label{subsec.Exp.Setup&TaskDesign}
\begin{figure}[t]
    \centering
    \includegraphics[width=0.7\columnwidth]{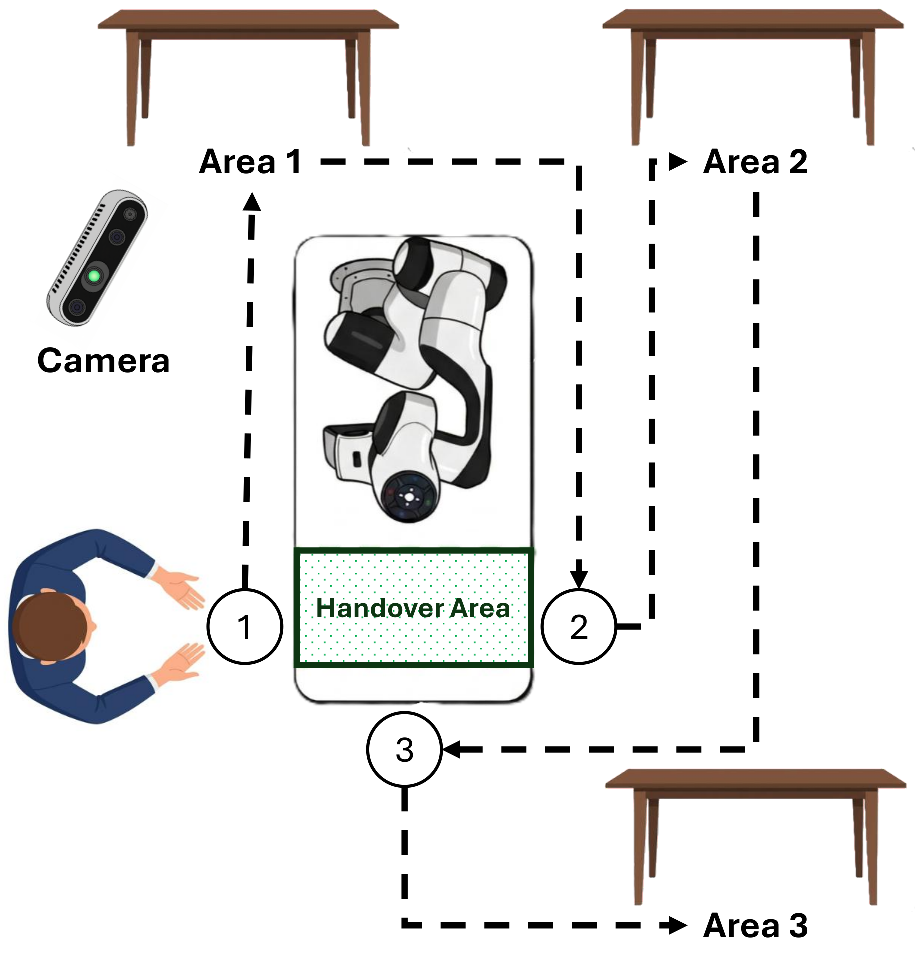}
    \caption{Overview of the laboratory experimental setup, highlighting the robotic arm, the vision system, and the predefined handover and deposition areas.}
    \label{fig:lab_setup}
\end{figure}

Building upon the hardware architecture previously described in \ref{subsec.Apparatus}, the experimental workspace is systematically organized to facilitate repeatable human-robot interactions. As illustrated in Fig. \ref{fig:lab_setup}, the layout features three predefined handover spots where the user stands to receive the objects, paired with three corresponding deposition areas for placing the items post-handover. A predefined handover volume is established within the robot's workspace, where users are instructed to place their right hand. The Intel RealSense camera continuously monitors this zone to enable real-time estimation of the hand pose.
The experimental trials involve two common household items: a mug and a smartphone. The handover tasks are evaluated under the two control modalities previously defined in \ref{subsec:control}: the baseline \textit{Static Handover} and the proposed \textit{Adaptive Handover}.
\subsection{Task Descriptions}
\begin{figure}[t] 
    \centering
    \includegraphics[width=0.30\columnwidth]{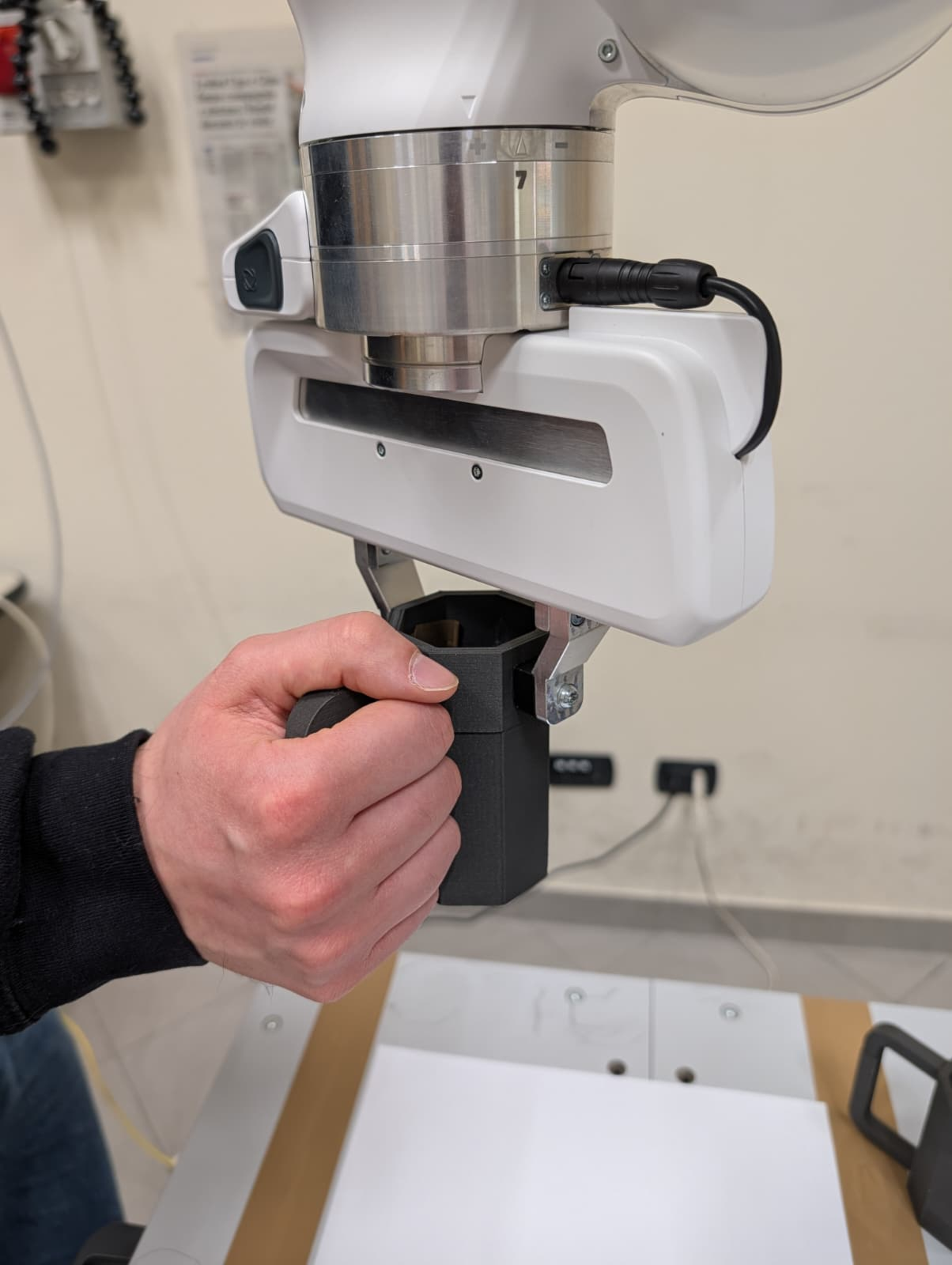}\hfill
    \includegraphics[width=0.30\columnwidth]{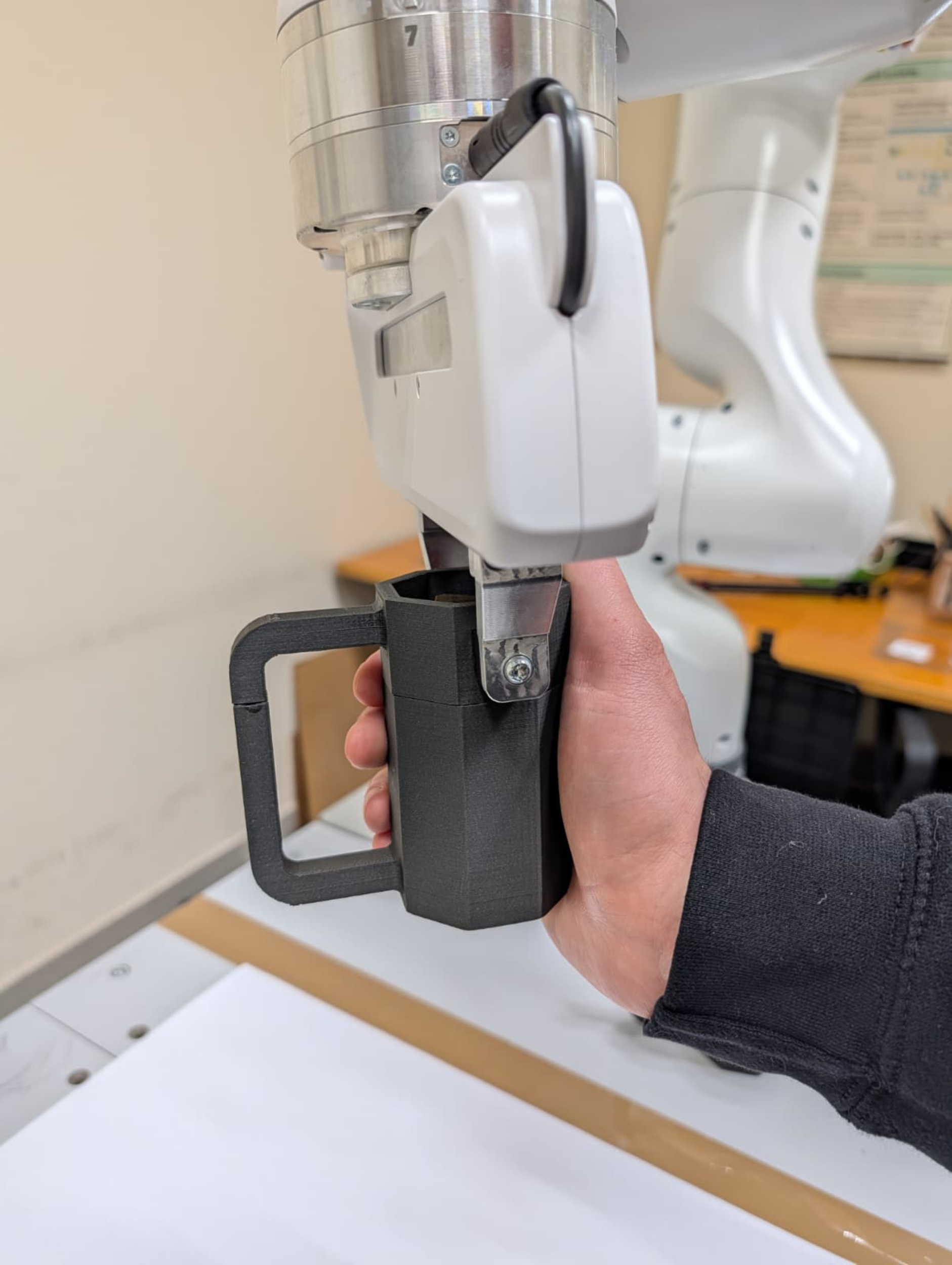}\hfill
    \includegraphics[width=0.30\columnwidth]{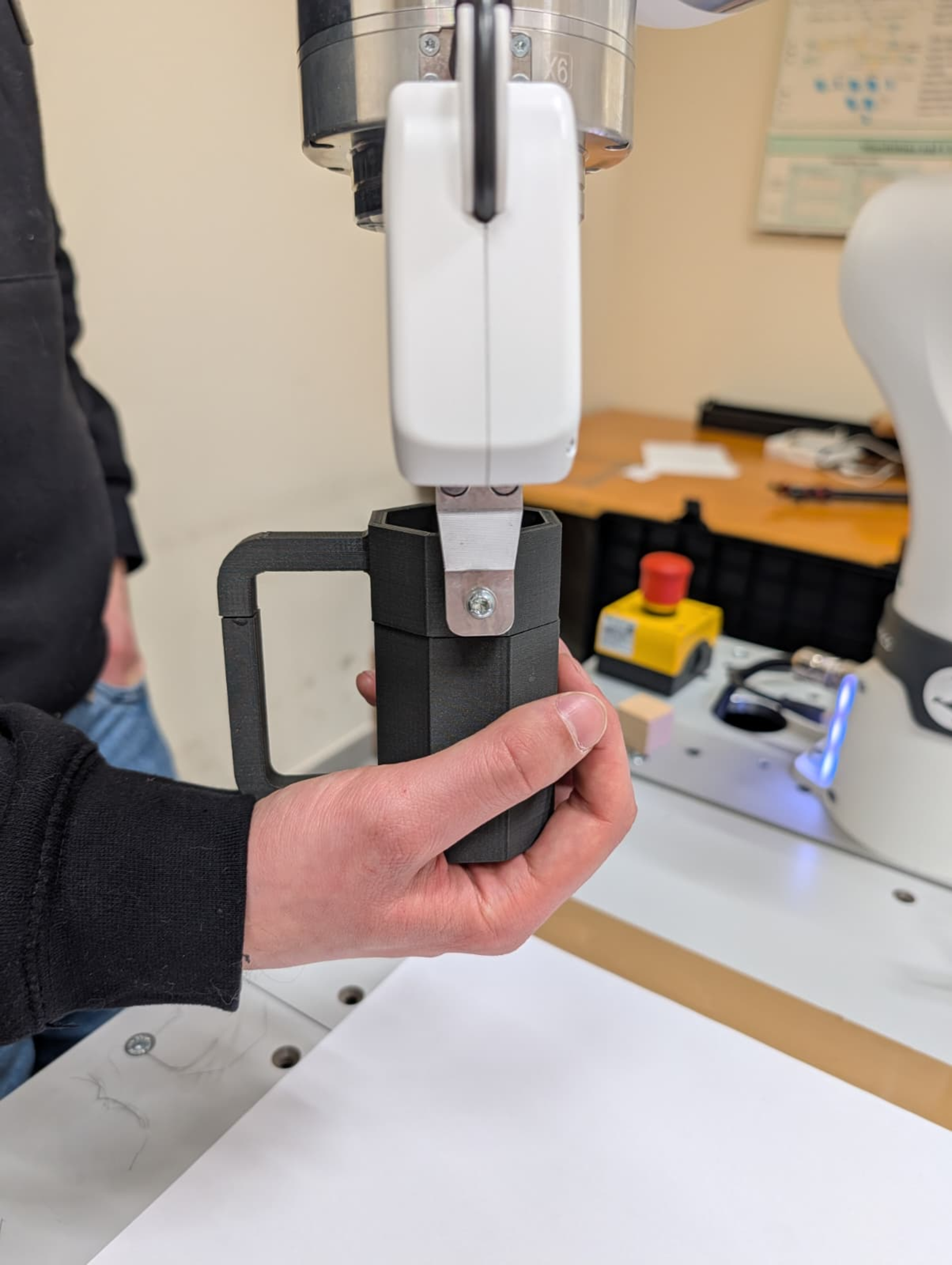}

    \vspace{0.5ex} 

    \includegraphics[width=0.30\columnwidth]{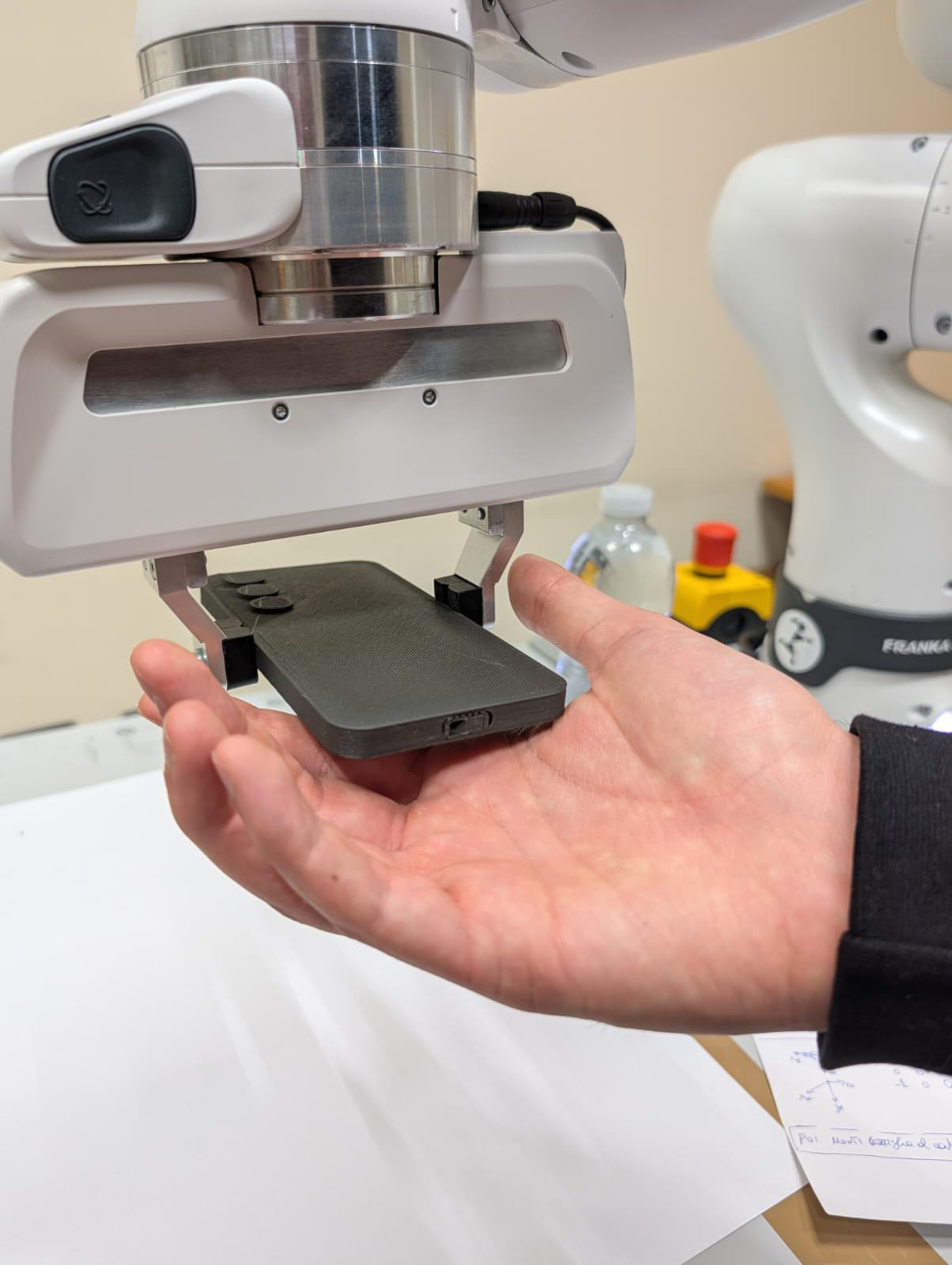}\hfill
    \includegraphics[width=0.30\columnwidth]{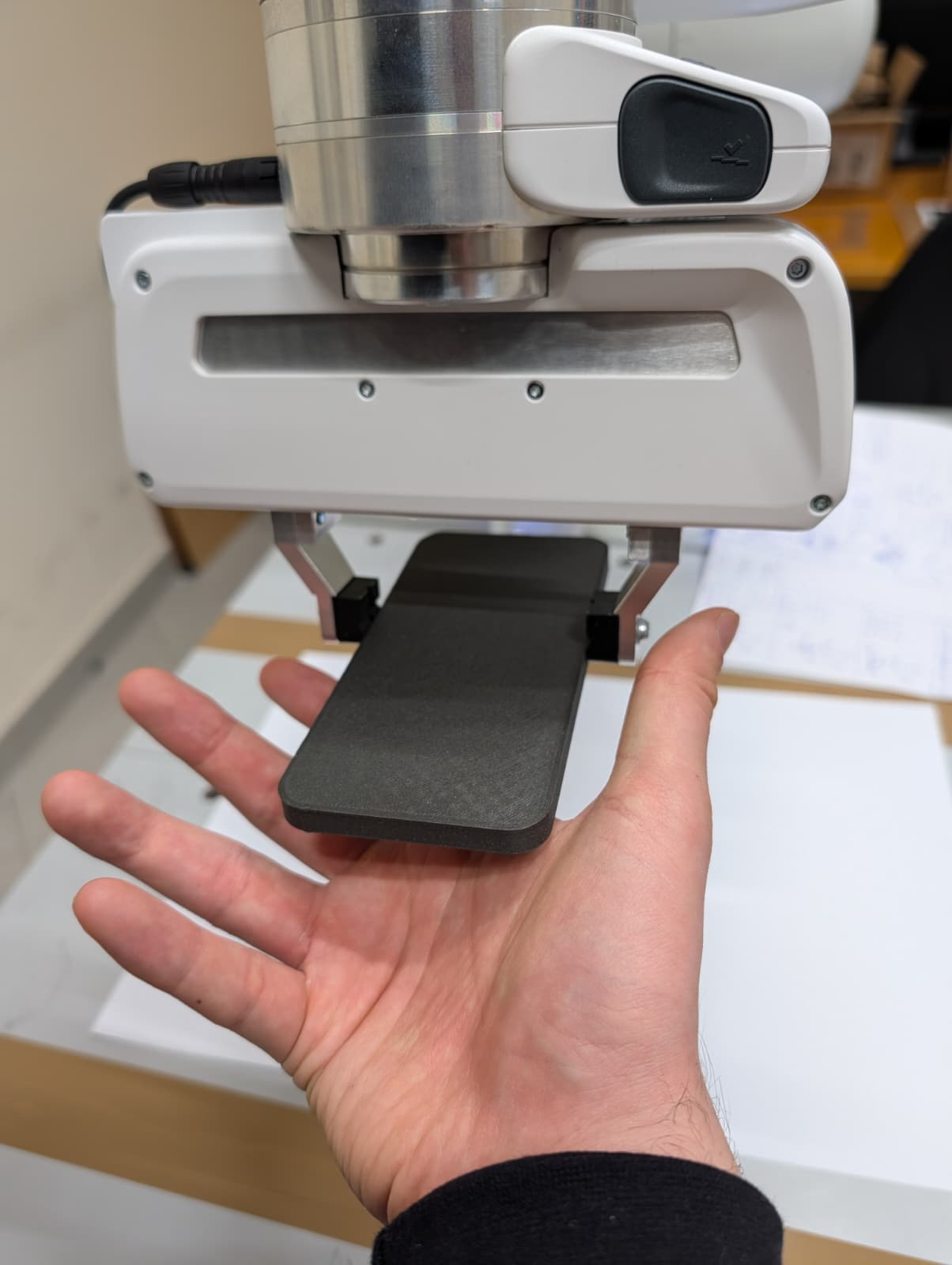}\hfill
    \includegraphics[width=0.30\columnwidth]{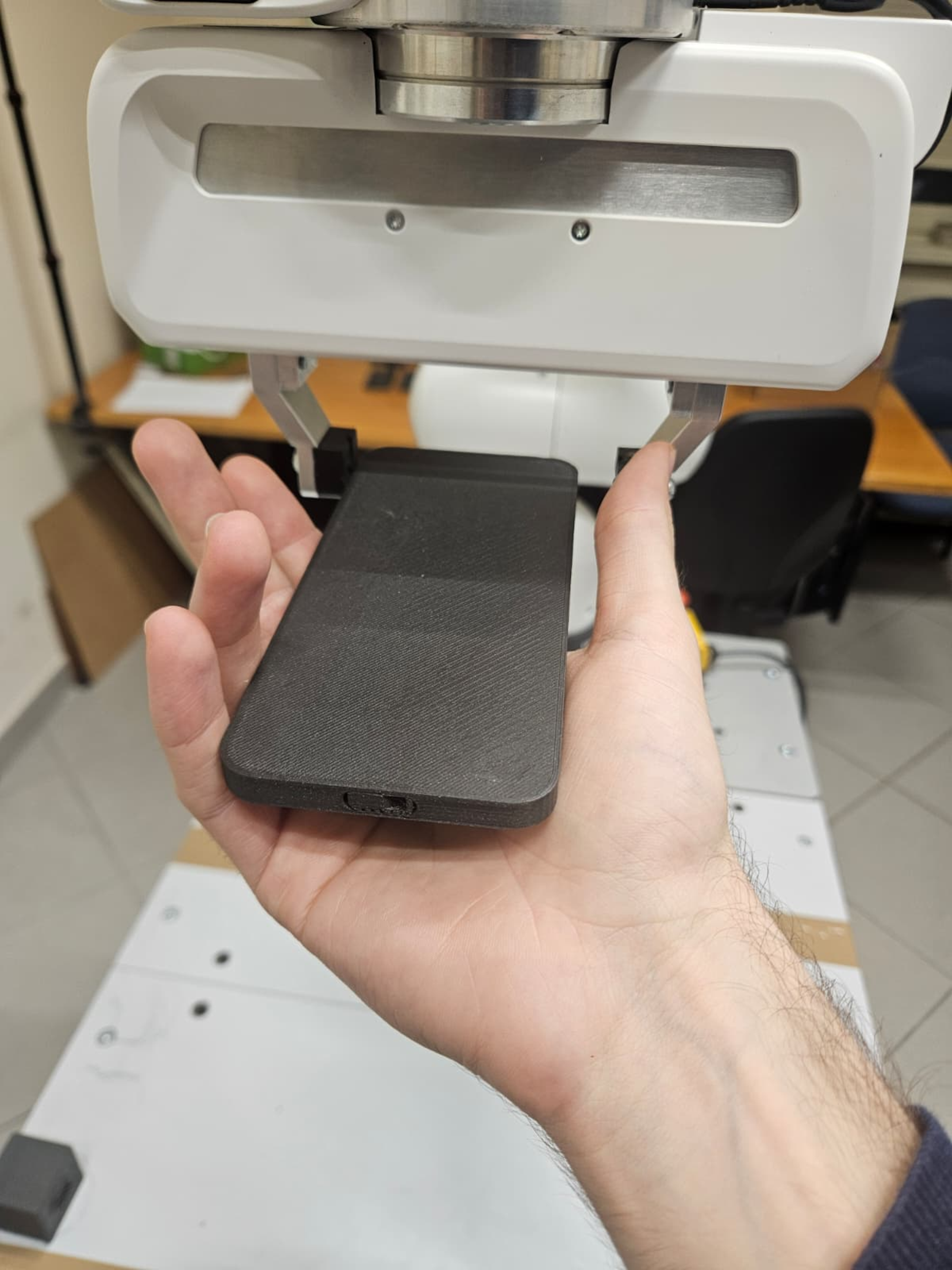}
    \caption{Visual overview of the task-dependent grasp types. \textit{Top row}: Mug orientations tailored for drinking (direct handle grasp), passing, and loading a dishwasher (bottom grasp). \textit{Bottom row}: Smartphone grips intended for placing, passing, and charging. The robot's adaptive handover alignment ensures ergonomic affordance for each specific post-handover task.}
    \label{fig:object_grasps_grid}
    
\end{figure}

To evaluate the system's adaptability, users are expected to receive the objects according to the task-specific grasps depicted in Fig. \ref{fig:object_grasps_grid} and execute the corresponding post-handover actions:
\\
\noindent \textit{Mug Tasks:} 
\begin{enumerate}
    \item \textit{Drinking:} The user receives the mug intending to drink from it, requiring a direct grasp on the handle.
    \item \textit{Passing:} The user receives the mug to hand it to another person, requiring a grasp that leaves the handle accessible to the receiver.
    \item \textit{Loading the Dishwasher:} The user grasps the mug from the bottom to place it upside down, simulating loading a dishwasher.
\end{enumerate}
\noindent \textit{Smartphone Tasks:}
\begin{enumerate}
    \item \textit{Placing:} The user receives the smartphone to lay it flat down on a table.
    \item \textit{Passing:} The user receives the smartphone to hand it to another person.
    \item \textit{Charging:} The user receives the smartphone with the intent to plug it into a charging cable.
\end{enumerate}
%
\begin{table}[t]
\centering
\caption{Ethical Acceptability Scale Items: Descriptive Statistics and Wilcoxon Test Results}
\label{tab:ethical_acceptability}
\renewcommand{\arraystretch}{1.2} 
\resizebox{\columnwidth}{!}{
\begin{tabular}{p{5.5cm}cccccc}
\toprule
\textbf{Ethical Acceptability Statement} & \textbf{Mean} & \textbf{SD} & \textbf{SE} & \textbf{CV} & \textbf{$V$} & \textbf{$p$} \\
\midrule
1. It is ethically acceptable cobots to be used to hand over objects to workers in industrial settings.
& 4.71 & 0.47 & 0.13 & 0.10 & 105.00 & < .001* \\
2. It is ethically acceptable for cobots to be used in domestic environments to assist people by handing them everyday objects.
& 4.43 & 0.65 & 0.17 & 0.15 & 91.00 & < .001* \\
3. It is ethically acceptable for the system to use cameras and artificial intelligence models to extract information from the environment.
& 3.64 & 1.22 & 0.33 & 0.33 & 44.00 & .047 \\
4. It is ethically acceptable for the introduction of these handover robots to partially replace human workers in logistics or assembly tasks.
& 4.14 & 0.66 & 0.18 & 0.16 & 78.00 & < .001* \\
5. It is ethically acceptable for these robots to reduce the need for human caregivers in basic domestic assistance.
& 4.14 & 1.10 & 0.29 & 0.27 & 84.00 & .003* \\
6. It is ethically acceptable for the human's pace of work or life to be somewhat influenced by the robot's reaction times.
& 3.15 & 1.21 & 0.33 & 0.39 & 26.50 & .334\\
7. It is ethically acceptable to delegate the user's physical safety to hardware limits, visual sensors, and robot control algorithms.
& 3.21 & 1.19 & 0.32 & 0.37 & 34.00 & .262 \\
8. It is ethically acceptable to program the robot's movements to appear deliberately ``human'' or ``natural'' in order to induce greater user trust.
& 4.07 & 0.92 & 0.25 & 0.22 & 74.00 & .003* \\
9. It is ethically acceptable to design or coat the robotic arm to give it a more friendly or organic appearance to promote acceptance.
& 3.93 & 0.92 & 0.25 & 0.23 & 62.00 & .004* \\
\bottomrule
\multicolumn{7}{p{12cm}}{\footnotesize \textit{Note.} $N=14$ (except for Item 6 where $N=13$). The One-Sample Wilcoxon signed-rank test evaluates the alternative hypothesis that the median acceptability score is greater than the neutral value of 3. * indicates statistical significance after Bonferroni correction.}
\end{tabular}
}
\vspace{-3mm}
\end{table}
\subsection{Procedure and Metrics}
\label{subsec:procedure_metrics}
The experiment followed a within-subjects design, where participants experienced both the Static and Adaptive handover conditions. To prevent ordering effects, the conditions were counterbalanced: half of the participants interacted with the Adaptive system first, while the other half started with the Static system. Within each condition, participants completed the same set of object-task trials. The object-task sequence was fixed across participants.
Before the interaction sessions, participants completed two baseline assessments. First, the 10-item Edinburgh Handedness Inventory \cite{Oldfield_1971} was administered to formally confirm their right-hand preference. Second, participants completed a modified 9-item Ethical Acceptability Scale, originally developed by Peca et al. \cite{Peca_2016}, to characterize \textit{a priori} attitudes toward the technology. This questionnaire covered specific aspects of robotic integration, and participants were asked to indicate their level of agreement with each statement using a 5-point Likert scale ranging from 1 (``strongly disagree'') to 5 (``strongly agree''). Following these questionnaires, users received a safety briefing, were equipped with NEON Pupil Labs glasses wearable eye-trackers, and proceeded with the handover tasks.
At the beginning of each trial, participants were instructed to place their hand in the handover area according to the task they were asked to perform and to keep it stable for a short fixed interval before the final delivery pose was computed and the robot motion was triggered.
To provide a comprehensive evaluation, the study combined objective physiological measurements with subjective self-reported assessments:
\begin{itemize}
    \item \textit{Objective Physiological Metrics:} The NEON Pupil Labs glasses continuously recorded the user's eye movements during the handover execution. Specifically, we extracted the \textit{blink rate} (blinks/min). This visual parameter serves as a non-invasive, real-time indicator of the user's cognitive workload and level of trust towards the robot's motion predictability.
    \item \textit{Subjective Metrics:} At the end of each interaction session, participants were administered two validated questionnaires. The NASA-TLX was used to assess perceived workload. Participants first rated the task across six dimensions (mental demand, physical demand, temporal demand, effort, performance, and frustration) on a 20-point scale. Subsequently, they completed 15 pairwise comparisons to weight these dimensions and compute an overall workload score. Finally, a modified version of the Human-Robot Trust Scale was administered to evaluate the perceived reliability and safety of the interaction. Participants rated 10 statements (detailed subsequently in Section \ref{sec:results}, Table \ref{tab:trust_results}) on a 5-point Likert scale ranging from 1 (``strongly disagree'') to 5 (``strongly agree'').
\end{itemize}
%
\begin{table}[b]
\centering
\caption{Normality and Paired Samples T-Test for Workload and Physiological Metrics}
\label{tab:ttest_results}
\begin{tabular}{lcccc}
\toprule
\multirow{2}{*}{\textbf{Metric}} & \multicolumn{2}{c}{\textbf{Shapiro-Wilk ($W$)}} & \multicolumn{2}{c}{\textbf{Paired t-test}} \\
\cmidrule(lr){2-3} \cmidrule(lr){4-5}
& \textbf{Adaptive} & \textbf{Static} & \textbf{$t(13)$} & \textbf{$p$-value} \\
\midrule
Blink Rate & 0.88 & 0.90 & -2.73 & .02* \\
Weighted TLX & 0.93 & 0.93 & -3.19 & <.01* \\
\bottomrule
\multicolumn{5}{l}{\footnotesize \textit{Note.} * indicates statistical significance ($p \le .05$).}
\end{tabular}
\end{table}

\begin{table}[t]
\centering
\caption{Descriptive Statistics for the Human-Robot Trust Scale}
\label{tab:trust_results}
\renewcommand{\arraystretch}{1.2} 
\resizebox{\columnwidth}{!}{
\begin{tabular}{p{4.5cm}cccc}
\toprule
\multirow{2}{*}{\textbf{Trust Scale Item}} & \multicolumn{2}{c}{\textbf{Adaptive}} & \multicolumn{2}{c}{\textbf{Static}} \\
\cmidrule(lr){2-3} \cmidrule(lr){4-5}
& \textbf{Mean} & \textbf{SD} & \textbf{Mean} & \textbf{SD} \\
\midrule
\multicolumn{5}{l}{\textit{\textbf{Robot Motion and Pick-up Speed}}} \\
\midrule
The way the robot moved made me uncomfortable & 1.93 & 1.14 & 2.07 & 1.27 \\
I wasn't worried because the robot moved as I expected & 3.57 & 1.02 & 3.21 & 1.19 \\
The robot's speed made me uncomfortable & 1.79 & 0.89 & 1.93 & 0.92 \\
The speed at which the gripper picked up and released objects made me uneasy & 1.64 & 0.63 & 1.86 & 0.86 \\
\midrule
\multicolumn{5}{l}{\textit{\textbf{Robot and Gripper Reliability}}} \\
\midrule
I felt I could rely on the robot to perform its tasks & 3.71 & 0.91 & 3.00 & 0.88 \\
I knew the gripper wouldn't drop the objects & 3.86 & 0.86 & 2.93 & 1.07 \\
I felt confident cooperating with the robot & 3.79 & 0.89 & 3.43 & 1.02 \\
\midrule
\multicolumn{5}{l}{\textit{\textbf{Safe Co-operation}}} \\
\midrule
I felt safe interacting with the robot & 4.36 & 0.84 & 3.93 & 1.21 \\
I felt uncomfortable cooperating with the robot due to the complexity level of the task & 1.71 & 0.91 & 1.79 & 0.97 \\
If the task had been more complicated, I would have been more worried & 2.93 & 1.27 & 3.00 & 1.41 \\
\bottomrule
\multicolumn{5}{p{8cm}}{\footnotesize \textit{Note.} $N=14$. Responses recorded on a 5-point Likert scale.}
\end{tabular}
}
\end{table}
\section{Results and Discussion}
\label{sec:results}
To evaluate the two handover modes, we compared the Adaptive Handover against the Static Handover. Data collected from the $N=14$ participants were analyzed to assess \textit{a priori} ethical acceptance, physiological indicator related to interaction demand, subjective cognitive workload, and trust of the user. 

\subsection{Ethical Acceptability Scale}
Participants' responses to the Ethical Acceptability Scale were evaluated using one-sample Wilcoxon signed-rank tests against the neutral midpoint of 3. The results are summarized in Table \ref{tab:ethical_acceptability}, which reports the test statistic ($V$) and the corresponding $p$-value for each statement. Because nine separate tests were performed, a Bonferroni correction was applied to control the family-wise error rate. The significance threshold was therefore adjusted from $\alpha = .05$ to $\alpha = .006 (.05/9)$. 
Most items showed statistically significant positive deviations from the test value ($p < .006$ after Bonferroni correction), suggesting that participants generally rated these robotic applications as ethically acceptable above the midpoint. Significant effects were observed for the use of cobots in industrial and domestic object handovers, the partial replacement of human workers in logistics or assembly tasks, the reduction of reliance on human caregivers, the use of human-like robot movements to foster trust, and the adoption of more friendly or natural robot appearances.
In contrast, the scores did not differ significantly from the neutral value for items involving AI-driven environmental data extraction ($p = .047$), robot-imposed pacing constraints ($p = .334$) and the delegation of physical safety to the hardware ($p = .262$).
Overall, the results suggest that participants held a generally favorable attitude toward robotic handover applications, while expressing concerns regarding data extraction, safety delegation, and potential constraints imposed by robotic timing.
\subsection{Physiological Stress and Workload}
\begin{figure}[h!]
    \centering
    \includegraphics[width=\columnwidth]{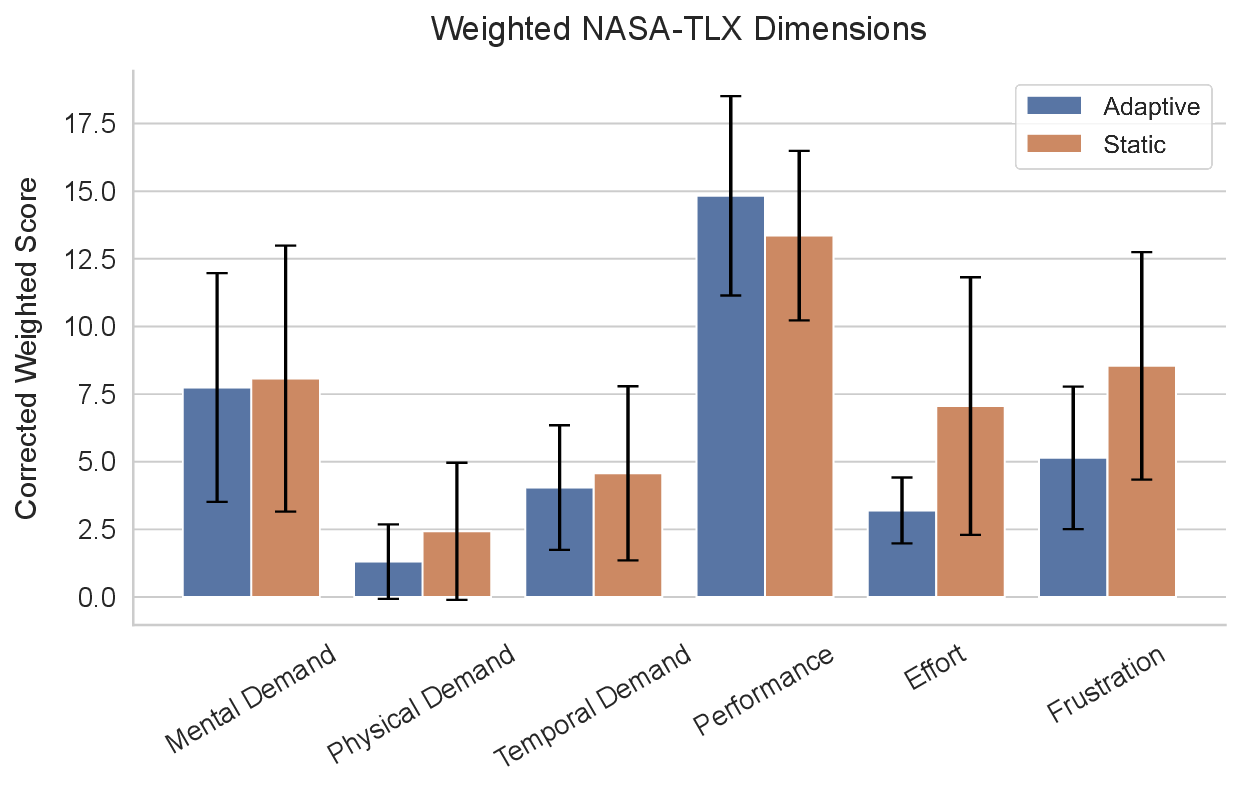}
    \caption{Comparison of the corrected weighted scores across the six NASA-TLX dimensions.}
    \label{fig:tlx_dimensions_bar}
    \vspace{-3mm}
\end{figure}

\begin{figure}[tb]
    \centering
    \includegraphics[width=0.9\columnwidth]{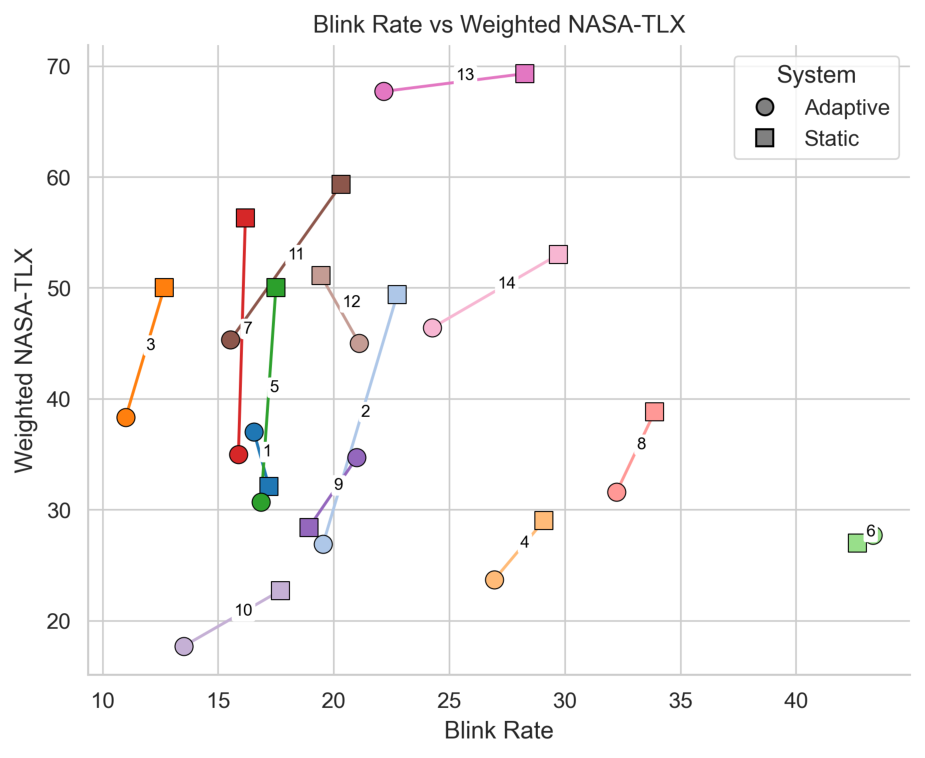}
    \caption{Participant-level paired scatter plot of blink rate versus weighted NASA-TLX. Each color corresponds to one participant. Circular markers denote the Adaptive Handover system, square markers denote the Static system, and the connecting lines link repeated observations within participant.}
    \label{fig:blink_tlx_scatter}
    \vspace{-3mm}
\end{figure}

Following Shapiro-Wilk tests confirming normality ($p > .05$), paired-samples t-tests were used to compare conditions (see Table~\ref{tab:ttest_results}). Real-time eye-tracking revealed significantly lower blink rates during the Adaptive Handover compared to the Static baseline. Consistent with psychophysiological literature \cite{Ifrah_2025}, this reduction provides supportive physiological
evidence of reduced interaction stress in the Adaptive condition. A similar pattern was observed in the subjective workload ratings. The Static system elicited a significantly higher weighted NASA-TLX score than the Adaptive system (Table~\ref{tab:ttest_results}). This indicates that the non-adaptive system imposed greater task demand and strain, likely requiring cumbersome postural adjustments or regrasping. A breakdown across the six NASA-TLX dimensions (Fig.~\ref{fig:tlx_dimensions_bar}) highlights a consistent trend of reduced demand with the Adaptive framework, despite expected interpersonal variability.
Fig.~\ref{fig:blink_tlx_scatter} presents a paired scatter plot at the participant-level, showing that the Static system was generally associated with a higher perceived workload and, in several cases, a higher blink rate compared to the Adaptive one.
\subsection{Human-Robot Trust Scale}
Questionnaire items were grouped according to the theoretical dimensions identified in the original scale: \textit{Robot Motion and Pick-up Speed}, \textit{Robot and Gripper Reliability}, and \textit{Safe Co-operation}. Descriptive statistics for the two systems are reported in Table~\ref{tab:trust_results}. Wilcoxon signed-rank tests were conducted to compare ratings between the two systems. To control for multiple comparisons, Bonferroni corrections were applied separately within each dimension.
Within the \textit{Robot Motion and Pick-up Speed} dimension, no significant differences were observed between systems for discomfort related to the robot's movement ($W = 25$, $z = -0.26$, $p = .836$), predictability of movement ($W = 37$, $z = 0.97$, $p = .331$), discomfort related to the robot's speed ($W = 7$, $z = -0.73$, $p = .484$), or uneasiness regarding the gripper's pick-up and release speed ($W = 8$, $z = -1.01$, $p = .299$).
Within the \textit{Robot and Gripper Reliability} dimension, participants reported significantly different ratings for the item assessing confidence that the gripper would not drop objects ($W = 36$, $z = 2.52$, $p = .013$). This effect remained significant following Bonferroni correction for the three items within this dimension (adjusted $\alpha = .017$). No significant differences were found for perceived robot reliability ($W = 46$, $z = 1.89$, $p = .052$) or confidence in cooperating with the robot ($W = 24$, $z = 0.84$, $p = .429$).
Within the \textit{Safe Co-operation} dimension, no significant differences were observed for perceived safety during interaction ($W = 11$, $z = 0.94$, $p = .416$), discomfort related to task complexity ($W = 6.5$, $z = -0.27$, $p = .892$), or anticipated concern under increased task complexity ($W = 9$, $z = -0.31$, $p = .824$).
Overall, the two systems were perceived similarly across most trust-related dimensions. The only significant difference was observed for participants' confidence that the gripper would securely retain objects during task execution in the Adaptive mode, suggesting a difference in perceived gripper reliability between the two systems. 
\section{Conclusion}
\label{sec:conclusion}
This study presents an adaptive robot-to-human handover framework that dynamically aligns object affordances to the user's hand and intended downstream task. Experimental validation, grounded in objective physiological metrics and subjective questionnaires, demonstrated the promising potential of this approach over a traditional static baseline. Specifically, dynamic alignment effectively mitigated the users' mental workload and physiological stress, as evidenced by the significantly lower NASA-TLX scores and decreased blink rate. With respect to trust, the two systems were perceived similarly across most dimensions, with the adaptive system yielding significantly higher confidence only in the gripper's ability to securely retain objects. Despite these encouraging results, several limitations exist. Beyond a small sample size ($N=14$) and limited object categories, the evaluation required participants to keep hands approximately stationary once the delivery pose was detected. Furthermore, the controlled static baseline may not fully capture the complexities of unconstrained collaboration. Finally, technical limitations remain regarding the AI vision models' inference speed and the camera's depth-estimation precision, which proved sensitive to the viewing angle and environmental occlusion. Nevertheless, the statistical trends observed across both physiological and subjective metrics indicate a robust underlying effect, providing a solid foundation for future experimental validation. Future work will focus on two main technological advances. First, we aim to enhance the system's autonomy by employing algorithms capable of automatically inferring the optimal grasping pose in real-time based on the recognized object, completely generalizing the approach to novel objects regardless of how the robot initially grasps them. Second, we plan to replace the standard Franka parallel gripper with a fully articulated anthropomorphic robotic hand. This hardware upgrade will allow us to investigate how a more human-like embodiment influences the user's psychological perception, trust dynamics, and overall acceptance during handover tasks.

\bibliographystyle{IEEEtran}
\bibliography{references}

\end{document}